\documentclass[10pt,twocolumn,letterpaper]{article}

\usepackage{cvpr}
\usepackage{times}
\usepackage{epsfig}
\usepackage{graphicx}
\usepackage{amsmath}
\usepackage{amssymb}
\usepackage{multirow}
\usepackage{booktabs}
\usepackage{makecell}
\usepackage{url}
\begin{document}

\title{ TriGait: Aligning and Fusing Skeleton and Silhouette\\
Gait Data via a Tri-Branch Network
}

\author{Yan Sun$^1$, Xueling Feng$^1$, Liyan Ma$^1$, Long Hu$^1$, Mark Nixon$^2$\\
$^1$School of Computer Engineering and Science, Shanghai University, China\\
$^2$School of Electronics and Computer Science, University of Southampton, United Kingdom\\
{\tt\small \{yansun, fengxueling, liyanma, longhu\}@shu.edu.cn, msn@ecs.soton.ac.uk}
}

\maketitle
\thispagestyle{empty}

\begin{abstract}
   Gait recognition is a promising biometric technology for identification due to its non-invasiveness and long-distance. However, external variations such as clothing changes and viewpoint differences pose significant challenges to gait recognition. Silhouette-based methods preserve body shape but neglect internal structure information, while skeleton-based methods preserve structure information but omit appearance. To fully exploit the complementary nature of the two modalities, a novel triple branch gait recognition framework, TriGait, is proposed in this paper. It effectively integrates features from the skeleton and silhouette data in a hybrid fusion manner, including a two-stream network to extract static and motion features from appearance, a simple yet effective module named JSA-TC to capture dependencies between all joints, and a third branch for cross-modal learning by aligning and fusing low-level features of two modalities. Experimental results demonstrate the superiority and effectiveness of TriGait for gait recognition. The proposed method achieves a mean rank-1 accuracy of 96.0\% over all conditions on CASIA-B dataset and 94.3\% accuracy for CL, significantly outperforming all the state-of-the-art methods. The source code will be available at \url{https://github.com/feng-xueling/TriGait/}.
\end{abstract}

\section{Introduction}

Gait recognition captures unique static and dynamic features from walking patterns of human for identification. Compared to other biometric technologies, gait can be recognized at a long distance, does not require active cooperation from the subject, and is difficult to disguise and easily change. Thus, it has great application prospects in many fields, e.g., public security and crime investigation. With the emergence of deep learning, gait recognition has made significant progress in the last decades~\cite{connor2018biometric}. However, external variations such as carrying objects, wearing different clothing, and changes in viewpoints have brought significant challenges to gait recognition. 

\begin{figure}[t]
\vspace{+0.2cm}
\begin{center}
   \includegraphics[width=1.0\linewidth]{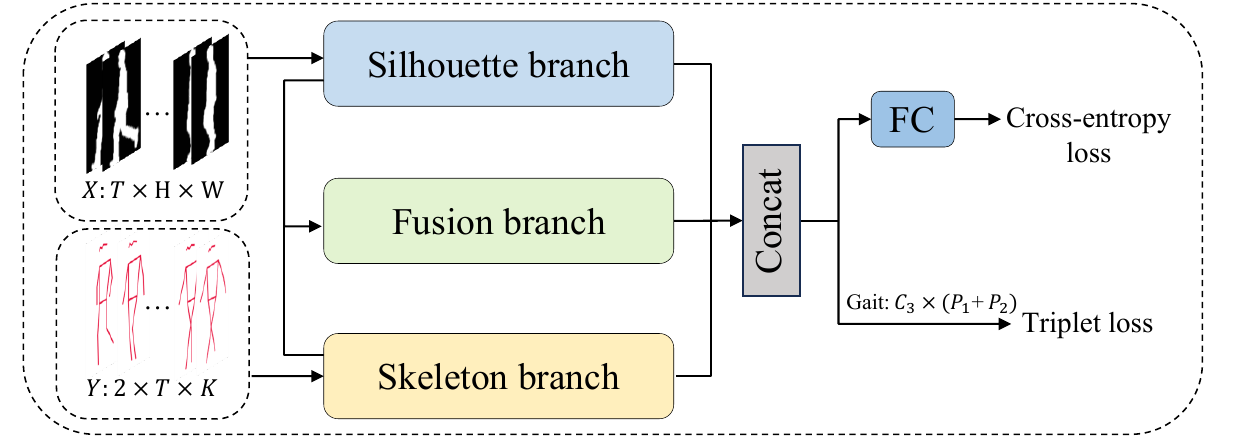}
\end{center}
   \caption{Overview of the framework TriGait. }
\label{fig:Overview}
\vspace{-0.2cm}
\end{figure}

To mitigate the influence of covariates, methods based on multiple data modalities have been proposed, e.g., silhouette~\cite{li2020gait, lin2021gait}, skeletons~\cite{liao2020model,mao2020gait}, depth images~\cite{sivapalan2011gait} and optical flow~\cite{castro2017evaluation}. Modalities such as depth and optical flow enrich the gait information~\cite{connor2018biometric}. However, obtaining such data is inconvenient and difficult to deploy into outdoor systems. Silhouette-based and skeleton-based methods are the most popular data in current research for gait recognition. Silhouette-based methods extract gait features from silhouette sequences, eliminating the effects of lighting, clothing texture, and color variations. However, the silhouette-based approach may fail to capture certain body structure information, especially when arms and torso overlap during walking. Furthermore, such methods are highly sensitive to external variations, such as carrying objects, changes in clothing, and variations in camera viewpoint, as they can significantly alter human appearance. The skeleton-based approach utilizes human pose estimation~\cite{sun2019deep} to extract the coordinates of human joints during walking. While the skeleton accurately preserves the structural information of the human body and mitigates the impact of external factors, such as carrying conditions and varying clothing, it tends to lose essential distinguishing appearance information.

\begin{figure*}
\begin{center}
\includegraphics[width=0.9\linewidth]{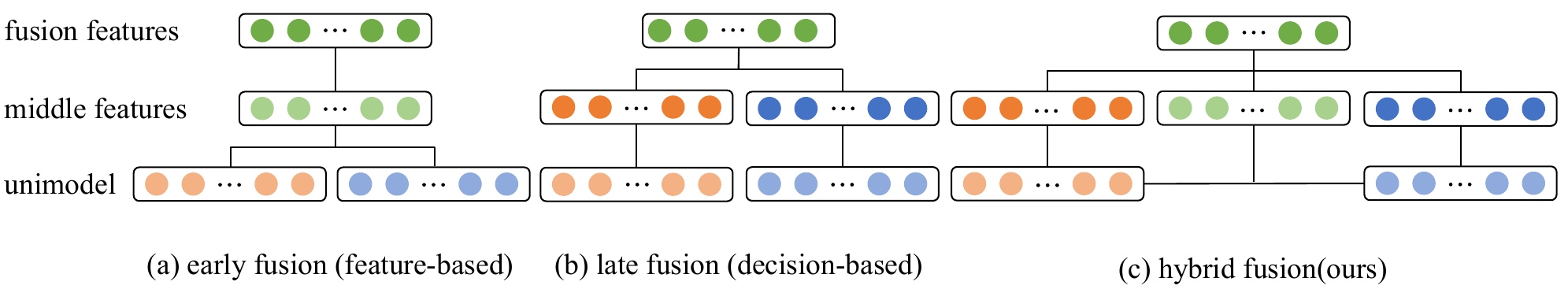}
\end{center}
   \caption{Three fusion types. (a) early fusion. (b) late fusion. (c) hybrid fusion.}
\label{fig:Fusionway}
\end{figure*}

Multimodal data provides complementary information that may not be available in a single modality, which has been widely used in the field of computer vision. Lan et al.~\cite{baltruvsaitis2018multimodal} divide the multimodal fusion approach into three types as illustrated in Figure~\ref{fig:Fusionway}. Since the skeleton is composed of discrete coordinate points while the silhouettes are sets of points most of the existing gait recognition methods utilize, the heterogeneity of the data leads to the difficulty to construct early fusion networks, as shown in Figure~\ref{fig:Fusionway}a. Besides, the early fusion is insufficient to fully utilize the high-level semantic information of the single modality. To the best of our knowledge, most of the existing gait recognition methods~\cite{peng2021learning, wang2023combining} utilize a post-fusion (decision-based) strategy, where unimodal gait features are extracted separately by two single branches, and then are combined using fusion algorithms such as concatenation and summation, as illustrated in Figure~\ref{fig:Fusionway}b. Although this approach can train on each modality better and more flexibly, it ignores the low-level interaction between the modalities in early fusion. Hybrid fusion attempts to exploit the advantages of both of the described methods in a common framework, as shown in Figure~\ref{fig:Fusionway}c, which has been used successfully for multimedia event detection~\cite{lan2014multimedia}.

To solve these problems, this paper proposes a novel triple-branch gait recognition framework to extract and fuse gait features from the two modalities using hybrid fusion, called TriGait, as show in Figure~\ref{fig:Overview}. Specifically, the two-stream network is adopted to extract the appearance and motion clues in silhouette. We apply multi-head self-attention in Transformer~\cite{vaswani2017attention} for the feature extraction of the skeleton. The Potential correlations between joints can be built to capture gait patterns. In addition to the unimodal branches, a third branch is proposed for aligning and fusing the low-level features from silhouette and skeleton, exploiting the complementary strengths of two modalities for effective gait recognition. Extensive experiments indicate that the proposed model outperforms all state-of-the-art methods on the popular CASIA-B dataset. The main contributions of the proposed method are summarized as follows:

\begin{itemize}
\item We propose a novel triple branch network (TriGait) that can effectively integrate features from the skeleton data and silhouette data in a hybrid fusion manner.
\item In the silhouette-based branch, the two-stream network is proposed to simultaneously extract appearance and motion features, which both play a significant role in gait recognition.
\item In the skeleton-based branch, we propose a simple yet effective joint self-attention module named JSA-TC to adaptively capture the dependencies between all nodes without considering the natural body structure. 
\item Experimental results show that the proposed method achieves superior gait recognition performance compared to previous state-of-the-art methods.  Extensive ablation studies validate the effectiveness of each proposed component of our method.
\end{itemize}

\section{Related Work}
\subsection{Silhouette-based Gait Recognition}

According to the temporal feature extraction method, the silhouette-based methods can be classified into two categories: templates-based and sequence-based. Templates-based methods~\cite{li2020gait} aggregate gait information into a single gait template, e.g., gait energy image (GEI) and gait entropy image (GEnI), and then extracts discriminative features from it. These methods are computationally efficient, but they do not take into account the temporal information and relationship between frames. Sequence-based methods~\cite{chao2019gaitset,fan2020gaitpart,lin2021gait} regard gait as a video sequence and employ the statistical functions or 3D-CNN-based model to extract temporal patterns. Chao et al.~\cite{chao2019gaitset} learned identity information from the set and aggregated convolutional maps across the entire sequence. Lin et al.~\cite{lin2021gait} construct spatiotemporal information in sequences with 3D CNNs.

Spatial gait feature representation learning can be categorized into global and local-based methods. Global-based methods~\cite{chao2019gaitset} extract holistic features directly from the silhouette, treating the human body as a whole. However, global representations are sensitive to occlusions and appearance changes due to the absence of body parts. In contrast, local-based approaches~\cite{fan2020gaitpart} divide the silhouette into local regions. Local features have been shown to enhance the robustness of gait recognition methods to occlusions. Combining both global and local representations has demonstrated better performance in gait recognition, as exemplified by the GaitGL~\cite{lin2021gait} method.

\subsection{Skeleton-based Gait Recognition}
The raw skeleton data predicted by pose estimation methods are the coordinates of key joints in the human body. Some skeleton-based gait recognition methods use hand-designed features based on joint point coordinates. PoseGait~\cite{liao2020model} defines three spatiotemporal pose features (i.e., joint angle, joint motion, and limb length) based on a priori knowledge and rearranged them to form a feature matrix as input. GCN has been widely used for action recognition because of its powerful ability to learn spatial patterns. There have been some works~\cite{mao2020gait,teepe2022towards} employed GCN for gait recognition as well. The human body is modeled as a topological map and graph convolutional networks (GCNs) are used to extract action features. GaitGraph~\cite{teepe2022towards} has made great progress in gait recognition by employing GCNs. However, due to the highly variable and nonlinear characteristics of gait data, the GCN may ignore some important joint features and fail to capture higher-order dependencies. There is still significant potential for improvement in skeleton-based gait recognition.

\subsection{Multimodal Gait Recognition}
To exploit the complementary nature of skeleton and silhouette, several bimodal fusion gait recognition methods have emerged. Wang et al.~\cite{wang2023combining} combine silhouette data and skeleton data for gait recognition based on late fusion. This method contains two branches, namely a CNN-based branch taking silhouettes as input and a GCN-based branch taking skeletons as input, which shows great potential for bimodal gait feature fusion. However, similar to most late fusion methods, it fails to capture low-level correlations of the two modalities, which may lead to underutilization of complementary information. TransGait~\cite{li2023transgait}, which employs an early fusion approach, converts the skeleton data into heatmaps, and then fuses the low-level features extracted from the skeleton heatmap and silhouette. Finally, it applies the transformer for temporal modeling. However, the skeleton heatmap requires higher computational and memory costs compared to the joint coordinates. Besides, it overlooks the high-level semantic information of the single modality.

\begin{figure*}
\begin{center}
   \includegraphics[width=0.7\linewidth]{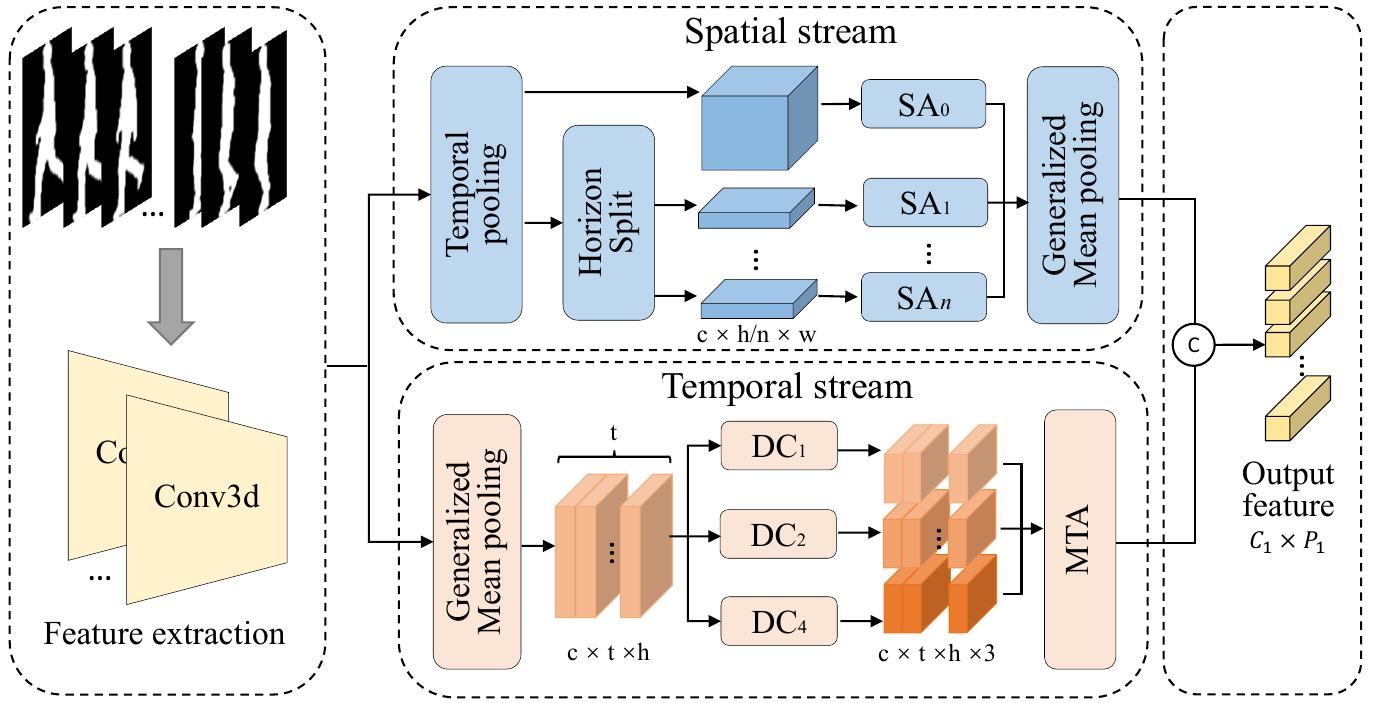}
\end{center}
   \caption{Structure of the two-stream silhouette branch. SA and MTA denote the spatial attention module in the spatial stream and the multi-scale temporal aggregation module in the temporal stream respectively.}
\label{fig:Silhouette}
\end{figure*}

Thus, we propose a novel triple-branch gait recognition framework to align gait features from the two modalities using hybrid fusion. This framework is able to extract unimodal features separately, as well as low-level interaction information between the two modalities.

\section{Method}
\subsection{Pipeline overview}
We propose a novel triple-branch neural network TriGait to exploit the complementary strengths of different data modalities for gait recognition. The overall pipeline is illustrated in Figure~\ref{fig:Overview}. TriGait integrates features from the skeleton and silhouette data in a hybrid fusion manner. Formally, given a gait sample with silhouette input $X$ and skeleton input $Y$, a two-stream network in silhouette branch is proposed to extract appearance and motion features $Gait_{sil}$ from $X$, a simple yet effective module named JSA-TC in skeleton branch is applied to capture high-order dependency between joints $Gait_{ske}$ from $Y$. For the Fusion branch, we design a cross-modal learning method to align and combine the low-level features extracted from the two modalities, denoted as $Gait_{fuse}$. The proposed network obtains the multimodal gait features $Gait$ from the bimodal fusion of $X$ and $Y$ through the steps in Eq.~\ref{pip1} to Eq.~\ref{pip3}:
\begin{equation}\label{pip1}
\begin{split}
Gait_{sil}=&\operatorname{Branch}_{sil}(X) \in \mathbb{R}^{C_{1} \times P_{1}}\\
Gait_{ske}=&\operatorname{Branch}_{ske}(Y) \in \mathbb{R}^{C_{2} \times P_{1}} \\
Gait_{fuse}=&\operatorname{Branch}_{fuse}(X, Y) \in \mathbb{R}^{C_{3} \times P_{2}}
\end{split}
\end{equation}
\begin{equation} \label{pip2}
    Gait^{\prime}=\operatorname{FC}\left(\operatorname { Concat }\left({ Gait }_{{sil }}, \alpha \cdot { Gait }_{{ske }}\right)\right) \in \mathbb{R}^{C_{3} \times P_{2}}
\end{equation}
\begin{eqnarray} \label{pip3}
    Gait=  \operatorname{Concat}\left(Gait_{fuse}, Gait^{\prime}\right) \in \mathbb{R}^{C_{3} \times\left(P_{1}+P_{2}\right)}
\end{eqnarray}
where $C_i \times {P_i}$ denoteas the embedding dimension of each branch. Gait is generated by combining the unimodal gait features along the channel dimension with a learnable parameter $\alpha$ to balance the features scales from the two branches. Then, the fusion feature $Gait'$ is processed by a fully connected layer $\operatorname{FC(\cdot)}$ for dimensionality reduction. We will introduce the components in detail in the following parts.

\subsection{Silhouette-based Branch}
To explore the correlation between spatial and temporal, the input silhouette $X\in \mathbb{R}^{T\times 1\times H\times W}$, where $T$, $H$, $W$ represent the frames number, the height and width of each frame respectively, is first fed into the 3D convolution network to obtain motion features from adjacent frames while extracting appearance features, which are denoted as $F\in \mathbb{R}^{T\times C\times H\times W}$. Inspired by two-stream action recognition~\cite{simonyan2014two}, the gait feature $F$ is then divided into spatial stream and temporal stream, which are used to extract appearance features and motion features respectively, as shown in Figure~\ref{fig:Silhouette}.

In the \textbf{spatial stream}, temporal max-pooling is used to obtain global spatial feature $F_S\in \mathbb{R}^{C\times H\times W}$. To enhance the robustness to occlusions of the model and capture the most salient spatial information, the global features are divided into $n$ parts along the height dimension and apply spatial attention to each local and global part separately. Inspired by CBAM~\cite{woo2018cbam}, the spatial attention produces a spatial attention map $S_A$ to emphasize or suppress features in different spatial locations. As shown in Eq.~\ref{eq:sil01}, spatial feature $F_s$ is projected into a reduced dimension $\mathbb{R}^{C/r\times H\times W}$ using $1\times1$ convolution to integrate and compress the feature map across the channel dimension. After the reduction, two $3\times3$ dilated convolutions are applied to utilize contextual information effectively. Finally, the features are again reduced to $\mathbb{R}^{1\times H\times W}$ spatial attention map using $1\times1$ convolution. A sigmoid function is taken to obtain the final spatial attention map $S_A$ in the range from 0 to 1. This attention map is element-wisely multiplied with the input feature map $F_S$ then is added upon the input feature map to acquire the refined feature map $F_S'$. Formally, the spatial attention is computed as: 
\begin{equation}
    \begin{aligned}
        S_{A}=  \sigma\left(\operatorname{BN}\left(\operatorname{C}_{3}^{1 \times 1}\left(\operatorname{C}_{2}^{3 \times 3}\left(\operatorname{C}_{1}^{3 \times 3}\left(\operatorname{C}_{0}^{1 \times 1}\left(F_{S}\right)\right)\right)\right)\right)\right)
    \end{aligned}
    \label{eq:sil01}
\end{equation}
\begin{equation}
    \begin{aligned}
         F_{S}^{\prime}  =  F_{S}+F_{S} \otimes S_{A}
    \end{aligned}
    \label{eq:sil02}
\end{equation}
where $\operatorname{C}(\cdot)$ denote a convolution operation, $\operatorname{BN}(\cdot)$ denotes a batch normalization operation, the superscripts denote the convolutional filter sizes, $\sigma$ is a sigmoid function, $\otimes$ denotes element-wise multiplication. Finally, the global and local spatial attention features are fused and processed by GeM~\cite{lin2021gait} to adaptively integrate the final appearance feature $A \in \mathbb{R}^{C \times 1 \times H \times 1}$. The process can be defined as:
\begin{eqnarray}
\begin{aligned}
A=\operatorname{GeM}\left(G_{A}^{\prime} + \operatorname{Concat}\left(L_{A 1}^{\prime}, \ldots, L_{An}^{\prime}\right)\right) \\
\end{aligned}
\label{eq:sil03}
\end{eqnarray}
\begin{eqnarray}
\begin{aligned}
\operatorname{GeM}(\cdot)=\left(\operatorname{F}_{avg}^{1 \times 1 \times W}\left((\cdot)^{p}\right)\right)^{\frac{1}{p}}
\end{aligned}
\label{eq:sil04}
\end{eqnarray}
where $L_{A1}', \dots, L_{An}'$ represents $n$ local spatial features, $\operatorname{F}_{avg}$ denotes average pooling, $p$ is the parameter which can be learned by network training.

In the \textbf{temporal stream}, we proposed Multi-scale Temporal Aggregation (MTA) to capture short and long-term temporal relations. Firstly, GeM is employed to integrate the spatial information. As illustrated in Figure~\ref{fig:Silhouette}, three parallel dilated convolutions($\operatorname{DC}_i(\cdot)$) with dilation rates $i$ increasing progressively are used to cover various temporal ranges. We concatenate the outputs from three branches as the updated temporal feature $F_T \in \mathbb{R}^{C \times T \times H \times 3}$. The three features are then processed by MTA, which consists of a convolution block and a sigmoid function. By employing temporal self-attention to weight the importance of each frame from different scales and capture the contextual, temporal relations among inconsecutive frames can be captured by MTA. Formally, The final temporal attention map $S_T$ is obtained by:
\begin{eqnarray}
S_{T}=\sigma\left(\operatorname{Unflatten}\left(\operatorname{C}_{1}^{3 \times 3}\left(\operatorname{C}_{0}^{3 \times 3}\left(\operatorname{Flatten}\left(F_{S}\right)\right)\right)\right)\right.
\label{eq:sil05}
\end{eqnarray}
\begin{eqnarray}
F_{T}^{\prime}=F_{T}\otimes S_{T} 
\label{eq:sil06}
\end{eqnarray}
where c and $\operatorname{Flatten}(\cdot)$ represent the feature map reshaping, $F_T^{\prime} \in \mathbb{R}^{C \times T \times H \times 3}$. We employed the temporal pooling to aggregate the motion feature $M \in \mathbb{R}^{C \times 1 \times H \times 1}$. 

The final silhouette-based branch feature $Gait_{sil} \in \mathbb{R}^{2C \times H}$ is fused by appearance feature $A$ and motion feature $M$ using the $\operatorname{Concat}(\cdot)$ operation. By combining the spatial and temporal streams, our silhouette branch can effectively extract and integrate appearance and motion features from the input silhouette.
\begin{figure}[t]
\begin{center}
        \begin{subfigure}[t]{0.45\linewidth}
            \centering
            \includegraphics[width=\linewidth]{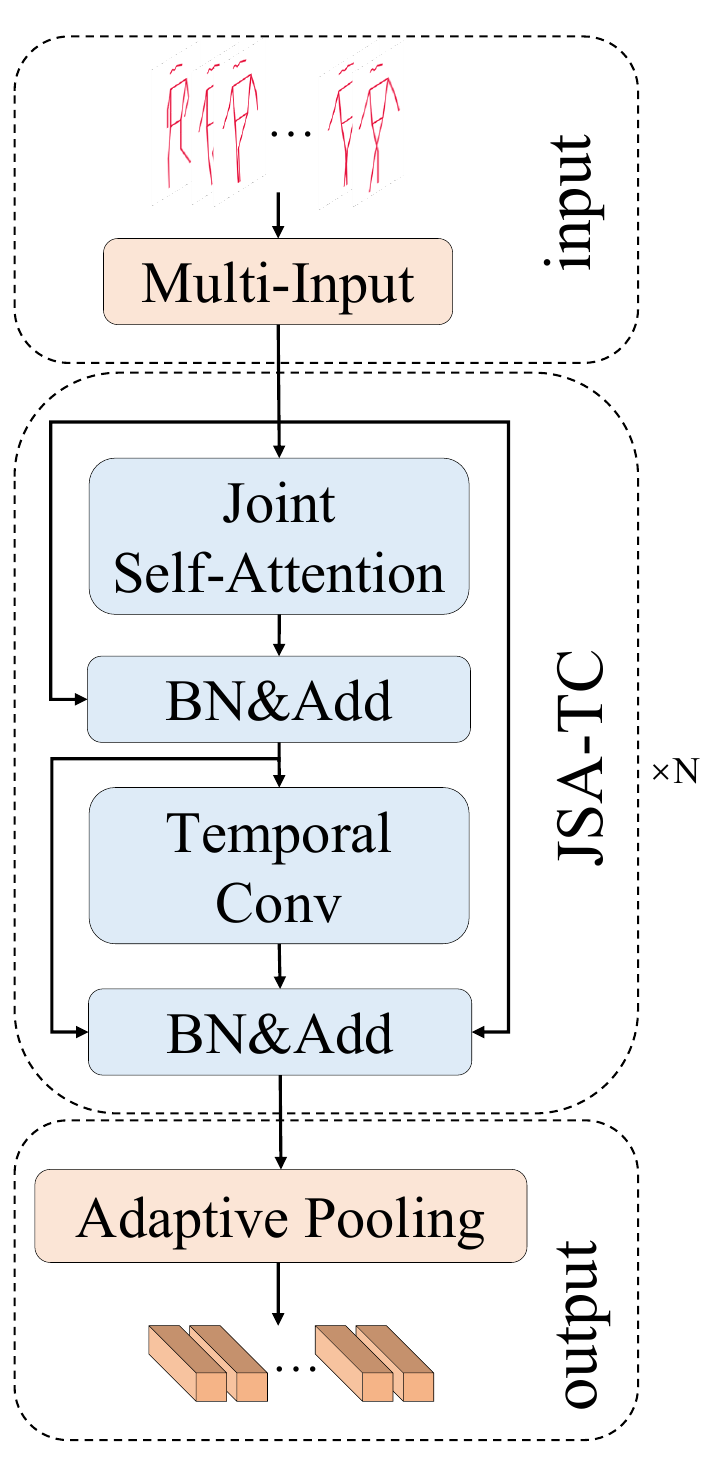}
            \caption{}
            \label{fig:JSA-TC}
        \end{subfigure}
        \hspace{0.4cm}
        \begin{subfigure}[t]{0.4\linewidth}
            \centering
            \includegraphics[width=\linewidth]{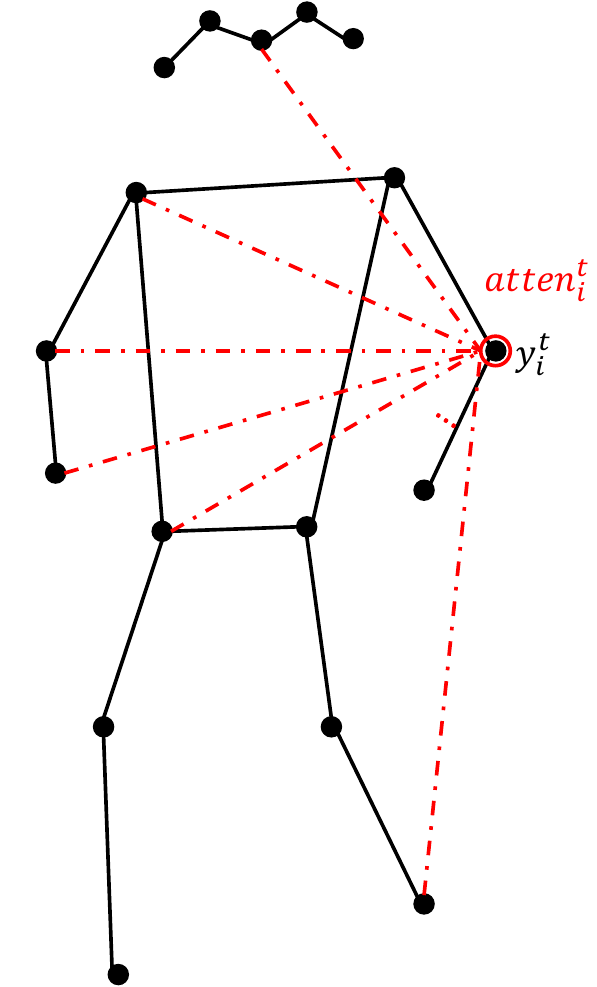}
            \caption{}
            \label{fig:JSA}
        \end{subfigure}
\end{center}
   \caption{The pipeline of skeleton branch. (a) Structure of JSA-TC module. (b) The visualization of joint self-attention (JSA).}
\label{fig:Skeleton}
\end{figure}

\subsection{Skeleton-based Branch}
Given the input of 2D skeleton $Y \in \mathbb{R}^{2 \times T \times K}$, $K$ denotes the number of joints on each frame and $2$ represents the channel number with coordinates $(x, y)$. Multi-Input (including joint, motion, and bone features) is implied in the skeleton branch which has been shown to be helpful in some human skeleton-based tasks~\cite{song2020stronger}. The three types of features are concatenated along the channel dimension and fed into the network.

In order to extract discriminative and robust descriptors from the skeleton data, we designed a new module called JSA-TC as shown in Figure~\ref{fig:JSA-TC}. In this module, joint self-attention (JSA) helps dynamically establish links between skeleton joints of each frame, while a temporal convolution (TC) is used to model the dynamic changes of each joint separately along all the frames. Besides, some residual connections are added in the module. 
JSA applies multi-head self-attention inside each frame to model correlations between each pair of joints in every single frame independently, as depicted in Figure~\ref{fig:JSA}. Given the $i$th joint feature $y_i^t$ at time t, a query vector $q_i^t\in\mathbb{R}^{d_k}$, a key vector $k_i^t\in\mathbb{R}^{d_k}$, and a value vector $v_i^t\in\mathbb{R}^{d_k}$ are first computed by applying three trainable linear transformations to the joint features. Then, for each pair of body joints $(i^t,j^t)$, a query-key dot product is applied to obtain a weight $\alpha_{ij}^t$ representing the strength of the correlations between the two nodes. The resulting score $\alpha_{ij}^t$ is used to weight each joint value $v_j^t$, and a weighted sum is computed as:
\begin{eqnarray}
\alpha_{i j}^{t}=q_{i}^{t} \cdot k_{j}^{t^{T}}, \forall t \in T
\end{eqnarray}
\begin{eqnarray}
    atten_{i}^{t}=\sum_{j} \operatorname{softmax}_{j}\left(\frac{\alpha_{i j}^{t}}{\sqrt{d_{k}}}\right) v_{j}^{t}
\end{eqnarray}
where $atten_i^t \in \mathbb{R}^{C_{\text{out}}}$ 
($C_out$ is the number of output channels) constitutes the spatial attention embedding of node $i_t$. The whole output of JSA $Atte\in \mathbb{R}^{C_{\text{out}} \times T \times K}$ applies multi-head self-attention ($head$=8) to compute spatial attention embeddings of each joint in each frame. The correlation structure is not fixed but it changes adaptively for each sample. 

Along the temporal dimension, the dynamics of each joint is studied separately along all the frames using TC. We employ a plain convolution network with the kernel size of 3 along the temporal dimension to extract motion clues in the skeleton. Lastly, we employed a temporal max pooling to extract the most discriminative features $f_{\text{ske}} \in \mathbb{R}^{C \times K}$
 across time. The adaptive pooling is employed to transform the spatial dimension of $f_{ske}$ from $K$ to $P_1$ for multi feature fusion as described in section 3.1, where $P_1$ equals to the $H$ in $Gait_{sil}$.

\begin{figure}[t]
\begin{center}
   \includegraphics[width=1.0\linewidth]{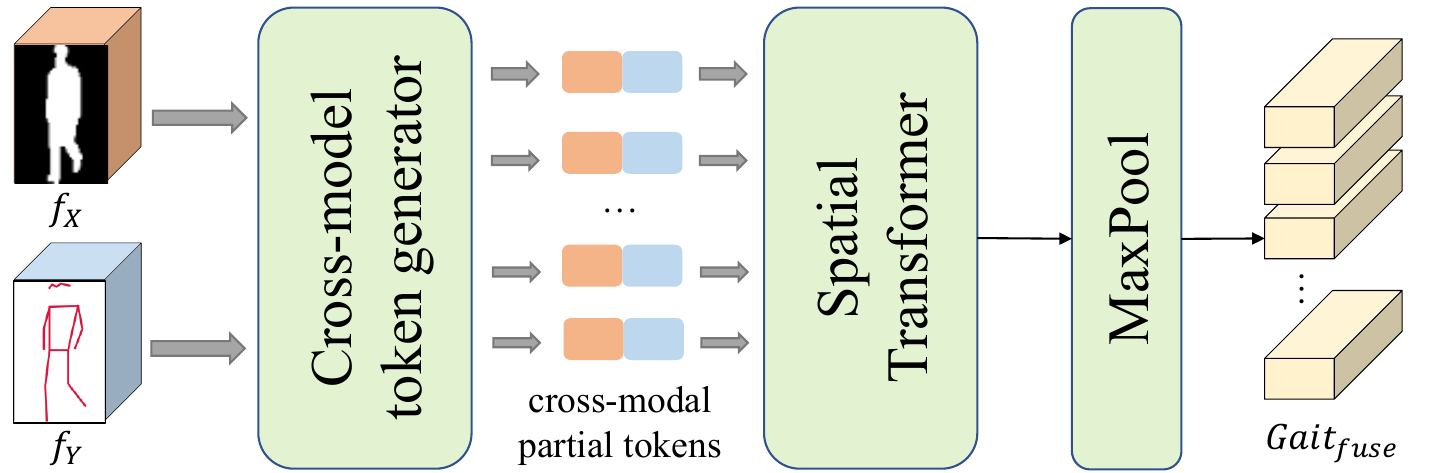}
\end{center}
   \caption{The pipeline of fusion branch for cross-modal learning. }
\label{fig:Fusion}
\end{figure}

\subsection{Fusion Branch}
Early fusion fails to capture the high-level semantic information of unimodality while late fusion can not exploit the low-level interaction between multi-modalities. Thus, Hybrid fusion is utilized in the work which can exploit the advantages of both described methods in a common framework. The skeleton is composed of discrete coordinate points while the silhouettes are grid data, which makes the early fusion a difficulty. To solve this, a cross-modal learning method is designed to align and combine the low-level features extracted from the two modalities as shown in Figure~\ref{fig:Fusion}. We propose a novel fusion mechanism called Cross-model token generator. In this generator, the human body is divided into 7 parts (e.g., head, shoulder, elbow, wrist, hip, knee, and ankle). For the skeleton, the division is easy to employ while the silhouette is the opposite. It is important to note that different individuals have distinct body proportions and walking patterns, which makes using a fixed partition a less optimal approach. Thus, we divide the silhouette feature horizontally based on the motion range of the upper 7 joint parts, enabling more precise alignment between the skeleton and silhouette data. Firstly, considering the inconsistent scale of skeleton coordinates across timestamps, we set the height of the neck to be 1 and scale all joints’ heights according to the scaling factor of each frame. After that, the flexion and extension position of each part $I^p$ is computed as:
\begin{eqnarray}
    H_{f}=\min \left(Y\left[:, x, I^{p}\right]\right) \\
H_{e}=\max \left(Y\left[:, x, I^{p}\right]\right)
\end{eqnarray}
where $x$ is the height of each joint, $H_f$ and $H_e$ are the flexion and extension horizontal position respectively, $I^p$ represents the related joints of the $p$th body part. The silhouette image was then partitioned into seven parts, allowing for overlapping, based on the displacement magnitude. Finally, max pooling and average pooling are used to extract the spatial information of each partitioned body part from the skeleton and silhouette. The fuse feature of the $p$th part can be obtained by:
\begin{eqnarray}
f_{X}^{p} & = & f_{X}\left[H_{f}: H_{e}\right] \\
f_{Y}^{p} & = & f_{Y}\left[I^{p}\right] \\
f_{X}^{p^{\prime}} & = & \operatorname{F}_{max}\left(f_{X}^{p}\right)+\operatorname{F}_{avg}\left(f_{X}^{p}\right) \\
f_{Y}^{p^{\prime}} & = & \operatorname{F}_{max}\left(f_{Y}^{p}\right)+\operatorname{F}_{avg}\left(f_{Y}^{p}\right) \\
P T_{fuse} & = & \operatorname{Concat}\left(f_{X}^{p^{\prime}}, f_{Y}^{p^{\prime}}\right)
\end{eqnarray}
where $f_{X} \in \mathbb{R}^{C_1 \times T \times H\times W}$, $f_{Y} \in \mathbb{R}^{C_2 \times T \times K}$ are the feature extracted from the first 3D convolution of silhouette branch and the first JSA-TC module of skeleton branch respectively, $PT_{fuse} \in \mathbb{R}^{(C_1+C2) \times T \times 7}$ are the aligned cross-modal partial tokens that contain the low-level interaction of two modalities.

Transformer~\cite{vaswani2017attention} is a powerful tool in terms of complexity correlation modeling. The self-attention mechanism in transformer is able to selectively attend to relevant features in both modalities, leveraging the latent correlation between the two modalities, so we utilize a spatial transformer self-attention module to extract the complementary information from the cross-modal partial tokens, as shown in Figure~\ref{fig:Fusion}. Due to the temporal misalignment of the two modalities, a plain max-pooling operation is applied along the temporal dimension to obtain the fusion representation $Gait_fuse$:
\begin{eqnarray}
Gait_{fuse} = \operatorname{F}_{max}\left(\operatorname{Transformer}\left(PT_{ fuse}\right)\right)
\end{eqnarray}
where $Gait_{fuse}\in \mathbb{R}^{2C\times7}$. The cross-modal learning method effectively integrates the complementary information provided by the two modalities.

\begin{table*}
 \renewcommand{\arraystretch}{1.038}
\begin{center}
\caption{The rank-1 accuracy (\%) on CASIA-B across different views, excluding the identical-view cases. Based on the walking condition, probe sequences are grouped into three subsets, i.e., NM, BG, and CL. TriGait stands for the proposed fusion network.}\label{tab:compare1}
\small 
\renewcommand\tabcolsep{5.0pt}
\begin{tabular}{l|l|lllllllllll|l}
\Xhline{1.5pt}
\multicolumn{2}{c|}{Gallery}                          & \multicolumn{11}{c|}{\begin{math}0^{o}-180^{o}\end{math}}                                               & \multirow{2}{*}{Mean} \\
\cline{1-13}
Probe               & \multicolumn{1}{c|}{Method}                        & {\begin{math}0^{o}\end{math}}   & {\begin{math}18^{o}\end{math}}  & {\begin{math}36^{o}\end{math}}  & {\begin{math}54^{o}\end{math}}  & {\begin{math}72^{o}\end{math}}  & {\begin{math}90^{o}\end{math}}  & {\begin{math}108^{o}\end{math}} & {\begin{math}126^{o}\end{math}} & {\begin{math}144^{o}\end{math}} & {\begin{math}162^{o}\end{math}} & {\begin{math}180^{o}\end{math}} &                       \\
\cline{1-14}
\multirow{8}{*}{NM} & GaitGraph~\cite{teepe2022towards} (CVPR2022)  & 78.5 & 82.9 & 85.8 & 85.6 & 83.1 & 81.5 & 84.3 & 83.2 & 84.2 & 81.6 & 71.8 & 82.0                  \\
                    & GaitMixer~\cite{pinyoanuntapong2022gaitmixer} (arXiv2022) & 94.4 & 94.9 & 94.6 & 96.3 & 95.3 & 96.3 & 95.3 & 94.7 & 95.3 & 94.7 & 92.2 & 94.9                  \\
                    \cline{2-14}
                    & GaitSet~\cite{chao2019gaitset} (AAAI2019)  & 90.8 & 97.9 & 99.4 & 96.9 & 93.6 & 91.7 & 95.0 & 97.8 & 98.9 & 96.8 & 85.8 & 95.0                  \\
                    & GaitPart~\cite{fan2020gaitpart} (CVPR2020)    & 94.1 & 98.6 & 99.3 & 98.5 & 94.0 & 92.3 & 95.9 & 98.4 & 99.2 & 97.8 & 90.4 & 96.2                  \\
                    & GaitGL~\cite{lin2022gaitgl} (arXiv2022)    & 96.6 & \textbf{98.8} & 99.1 & 98.1 & 97.0 & 96.8 & 97.9 & \textbf{99.2} & 99.3 & 98.3 & 95.6 & 98.0                  \\
                    & GaitMSTP~\cite{huang2022gaitmstp} (IJCB2022)    & \textbf{98.2} & 99.2 & 99.4 & 98.5 & 96.8 & 96.2 & 97.8 & 99.1 & 99.1 & \textbf{99.5} & 96.2 & \textbf{98.2}\\
                    \cline{2-14}
                    
                    & TransGait~\cite{li2023transgait}   (APPL INTELL2023)      &97.3 &99.6 &\textbf{99.7} &\textbf{99.0} &97.1 &95.4 &97.4 &99.1 &\textbf{99.6} &98.9 &95.8 &98.1                  \\
                    & Combine~\cite{wang2023combining} (ICASSP2023)      & 97.0 & 97.9 & 98.4 & 98.3 & 97.2 & \textbf{97.3} & 98.2 & 98.4 & 98.3 & 98.1 & 96.0 & 97.7                  \\
                    & TriGait (ours)                  & 97.0 & 98.6 & 98.3 & 98.3 & \textbf{98.4} & 97.0 & \textbf{98.6} & 99.0 & 98.9 & 98.4 & \textbf{97.4} & \textbf{98.2}                  \\
\cline{1-14}
\multirow{8}{*}{BG} & GaitGraph ~\cite{teepe2022towards} (CVPR2022)  & 69.9 & 75.9 & 78.1 & 79.3 & 71.4 & 71.7 & 74.3 & 76.2 & 73.2 & 73.4 & 61.7 & 73.2                  \\
                    & GaitMixer~\cite{pinyoanuntapong2022gaitmixer} (arXiv2022) & 83.5 & 85.6 & 88.1 & 89.7 & 85.2 & 87.4 & 84.0 & 84.7 & 84.6 & 87.0 & 81.4 & 85.6                  \\
                    \cline{2-14}
                    & GaitSet~\cite{chao2019gaitset} (AAAI2019)  & 88.3 & 91.2 & 91.8 & 88.8 & 83.3 & 91.0 & 84.1 & 90.0 & 92.2 & 94.4 & 79.0 & 87.2                  \\
                    & GaitPart~\cite{fan2020gaitpart} (CVPR2020)    & 89.1 & 94.8 & 96.7 & 95.1 & 88.3 & 84.9 & 89.0 & 93.5 & 96.1 & 93.8 & 85.8 & 91.5                  \\
                    & GaitGL~\cite{lin2022gaitgl} (arXiv2022)    & 93.9 & 97.3 & \textbf{97.6} & 96.2 & 94.7 & 91.0 & 94.4 & 97.2 & \textbf{98.6} & \textbf{97.1} & 91.6 & \textbf{95.4}                  \\
                    & GaitMSTP~\cite{huang2022gaitmstp} (IJCB2022)    & \textbf{95.7} & \textbf{97.5} & 97.4 & 96.1 & 93.4 & 90.8 & 93.4 & 96.4 & 98.1 & 96.9 & 92.6 & 95.3                  \\
                    \cline{2-14}
                    
                    & TransGait~\cite{li2023transgait}   (APPL INTELL2023)      &94.0 &97.1 &96.5 &96.0 &93.5 &91.5 &93.6 &95.9 &97.2 &\textbf{97.1} &91.6 &94.9\\
                    & Combine~\cite{wang2023combining}  (ICASSP2023)      & 91.9 & 94.6 & 96.4 & 94.3 & 94.4 & 91.6 & 94.1 & 95.4 & 95.5 & 93.9 & 89.5 & 93.8                  \\
                    & TriGait (ours)                  & 91.8 & 94.3 & 95.2 & \textbf{96.6} & \textbf{96.5} & \textbf{93.7} & \textbf{95.9} & \textbf{97.6} & 97.4 & 96.9 & \textbf{93.8} & \textbf{95.4}                  \\
\cline{1-14}
\multirow{8}{*}{CL} & GaitGraph~\cite{teepe2022towards} (CVPR2022)  & 57.1 & 61.1 & 68.9 & 66.0 & 67.8 & 65.4 & 68.1 & 67.2 & 63.7 & 63.6 & 50.4 & 63.6                  \\
                    & GaitMixer~\cite{pinyoanuntapong2022gaitmixer} (arXiv2022) & 81.2 & 83.6 & 82.3 & 83.5 & 84.5 & 84.8 & 86.9 & 88.9 & 87.0 & 85.7 & 81.6 & 84.5                  \\
                    \cline{2-14}
                    & GaitSet~\cite{chao2019gaitset} (AAAI2019)  & 61.4 & 75.4 & 80.7 & 77.3 & 72.1 & 70.1 & 71.5 & 73.5 & 73.5 & 68.4 & 50.0 & 70.4                  \\
                    & GaitPart~\cite{fan2020gaitpart} (CVPR2020)    & 70.7 & 85.5 & 86.9 & 83.3 & 77.1 & 72.5 & 76.9 & 82.2 & 83.8 & 80.2 & 66.5 & 78.7                  \\
                    & GaitGL~\cite{lin2022gaitgl} (arXiv2022)    & 82.6 & 92.6 & 94.2 & 91.8 & 86.1 & 81.3 & 87.2 & 90.2 & 90.9 & 88.5 & 75.4 & 87.3                  \\
                    & GaitMSTP~\cite{huang2022gaitmstp} (IJCB2022)    & 82.3 & 93.1 & 94.8 & 90.9 & 86.8 & 84.2 & 87.7 & 91.0 & 91.8 & 91.2 & 77.8 & 88.3                  \\
                    \cline{2-14}
                    & TransGait~\cite{li2023transgait} (APPL INTELL2023)      &80.1 &89.3 &91.0 &89.1 &84.7 &83.3 &85.6 &87.5 &88.2 &88.8 &76.6 &85.8\\
                    & Combine~\cite{wang2023combining} (ICASSP2023)      & 87.4 & \textbf{96.0} & \textbf{97.0} & 94.6 & 94.0 & 90.1 & 91.5 & 94.1 & 93.8 & 92.6 & 88.5 & 92.7                  \\
                    & TriGait (ours)                  & \textbf{91.7} & 93.2 & 96.9 & \textbf{97.0} & \textbf{95.2} & \textbf{94.0} & \textbf{94.6} & \textbf{95.3} & \textbf{94.1} & \textbf{94.1} & \textbf{90.8} & \textbf{94.3}                 \\
\Xhline{1.5pt}
\end{tabular}
\end{center}
\end{table*}

\begin{table*}
\vspace{+0.1cm}
\begin{center}

\caption{the rank 1 mean accuracy (\%) on CASIA-B across different conditions.}\label{tab:compare2}
\vspace{+0.2cm}

\begin{tabular}{l|l|lll|l}
\Xhline{1.5pt}
\multicolumn{1}{c|}{Input}         & \multicolumn{1}{c|}{Methods}           & NM   & BG   & CL   & Mean \\
\cline{1-6}
\multirow{2}{*}{Skeleton}               & GaitGraph~\cite{teepe2022towards}(CVPR2022)  & 82.0 & 73.2 & 63.6 & 72.9 \\
                                        & GaitMixer~\cite{pinyoanuntapong2022gaitmixer} (arXiv2022) & 94.9 & 85.6 & 84.5 & 88.3 \\
\cline{1-6}
\multirow{4}{*}{Silhouette}             & GaitSet~\cite{chao2019gaitset}(AAAI2019)   & 95.0 & 87.2 & 70.4 & 84.2 \\
                                        & GaitPart~\cite{fan2020gaitpart} (CVPR2020)   & 96.2 & 91.5 & 78.7 & 88.8 \\
                                        & GaitGL~\cite{lin2022gaitgl} (arXiv2022)   & 98.0 & \textbf{95.4} & 87.3 & 93.6 \\
                                        & GaitMSTP~\cite{huang2022gaitmstp} (IJCB2022)   & \textbf{98.2} & 95.3 & 88.3 & 93.9 \\
\cline{1-6}
\multirow{2}{*}{Multimodal } 
& TransGait~\cite{li2023transgait} (APPL INTELL2023)      & 98.1 & 94.9 & 85.8 & 92.9 \\
& Combine~\cite{wang2023combining} (ICASSP2023)      & 97.7 & 93.8 & 92.7 & 94.7 \\
                                        & TriGait (ours)                  & \textbf{98.2} & \textbf{95.4} & \textbf{94.3} & \textbf{96.0}\\

\Xhline{1.5pt}
\end{tabular}
\end{center}
\vspace{-0.3cm}
\end{table*}

\section{Experiment}
\subsection{Dataset}
\textbf{CASIA-B}~\cite{yu2006framework} is a popular dataset widely used in gait recognition. It contains 124 subjects and each subject has 11 views captured from 11 cameras simultaneously. The views are uniformly distributed in [$0^{o}$, $180^{o}$] at an interval of $18^{o}$. For each view, there are 10 sequences under three different conditions, i.e., 6 sequences in normal (NM), 2 sequences with a bag (BG), and 2 sequences in different clothes (CL). In total, there are 110 sequences per subject. The first 74 subjects are taken as the training set and the rest 50 as the testing set. For evaluation, the first 4 normal walking sequences of each subject are regarded as the gallery and the rest as the probe.
\subsection{Implementation Details}
All models are implemented with PyTorch and trained with NVIDIA 3090Ti GPUs.

\textbf{Input}. The silhouettes are pre-processed into aligned 64×64 images followed by~\cite{takemura2018multi}. Since the dataset does not provide skeleton data, the HRNet~\cite{sun2019deep} is applied in CASIA-B for 2D human pose estimation. The joint structure (with 17 key points) is followed by~\cite{teepe2022towards}. The number of subjects, the number of sequences per subject, and the number of frames per sequence in a mini-batch are set to (8, 16, 30).

\textbf{Network}. For the silhouette branch in TriGait, there are 4 3D convolution blocks with channels (1/32, 32/64, 64/128, 128/128). In the spatial stream, the global feature is divided into 8 parts. The reduction ratio r is set to 16. For the silhouette branch, there are 4 JSA-TC Blocks. We set the number of channels in the 4 blocks as (10/64, 64/64, 64/128, 128/256). The transformer layer is set to 1 in the fusion branch. The output dimension of each fused part feature is set to 256

\textbf{Training} The cross-entropy loss and Batch All (BA+) triplet loss~\cite{hermans2017defense} are applied to train the network. The margin m for triplet loss is set to 0.2. The loss $L$ for the TriGait is computed as:
\begin{eqnarray}\label{eq:loss}
L = L_{tri}+L_{ce}
\end{eqnarray}
where $L_{tri}$ and $L_{ce}$ donotes the triplet loss and cross-entropy loss repectively.
The model is trained over 60k iterations. The learning rate for the network is set to 1e-4 and is scaled to 1e-5 at 30K iterations. We use $momentum = 0.9$ and $weight decay = 5e-4$ for the optimization.

\textbf{Testing} In the testing phase, the probe sequences are grouped into three subsets, i.e., NM, BG, and CL. The evaluation is performed on each subset separately. All the results are averaged on gallery views excluding the identical-view cases.
\subsection{Comparison with State-of-the-Art Methods}
Table~\ref{tab:compare1} presents a comparison of state-of-the-art gait recognition methods on the CASIA-B dataset with the proposed TriGait fusion network. The accuracies of the compared approaches are directly cited from their original papers. The mean accuracy for all probes is shown in Table~\ref{tab:compare2}. The three blocks of each probe in Table~\ref{tab:compare1} and Table~\ref{tab:compare2} represent the state of arts of skeleton-based methods, silhouette-based methods, and multimodal methods respectively. The proposed TriGait model, which leverages the hybrid fusion of skeleton and silhouette, achieves outstanding performance on all probe sets, with a remarkable result of 98.2\% in NM, 95.4\% in BG, and 94.3\% in CL. The mean accuracy of TriGait achieves 96.0\% as shown in Table~\ref{tab:compare2}. As for unimodal methods, it can be observed that the accuracy of identification significantly decreases when there are large changes in appearance (e.g., in CL condition.). The GaitMSTP~\cite{huang2022gaitmstp} achieves the rank-1 accuracy at 98.2\% in NM but decreases to 88.3\% in CL, which means a unimodal approach (silhouette or skeleton) has inherent limitations under complex scenarios. Specifically, in the CL condition, all views on the proposed TriGait exceed 90\%, surpassing GaitMSTP by 6.0\% and Combine~\cite{wang2023combining} by 1.6\%. These results not only demonstrate the effectiveness and superiority of our approach but emphasize the significant potential of utilizing the complementary strengths of the skeleton and silhouette features to enhance the robustness of gait recognition against covariates.

\begin{table}
\begin{center}
\caption{Ablation experiments conducted for different network structures. The results are mean rank-1 accuracies over 11 views, with identical-view cases excluded.}\label{tab:ablation1}
\vspace{+0.2cm}

\begin{tabular}{ll|lll|l}
\Xhline{1.5pt}
\multicolumn{2}{c|}{Network   structure} & NM            & BG            & CL            & Mean          \\
\cline{1-6}
a         & Silhouette-branch           & 96.3          & 92.6          & 81.0          & 90.0          \\
b         & Skeleton-branch             & 92.3          & 84.4          & 79.9          & 85.5          \\
c         & Two-branch                  & 97.4          & 94.4          & 90.3          & 94.0          \\
d         & Tribranch (with UP)         & 97.7          & \textbf{95.7} & 92.9          & 95.4          \\
e         & TriGait (ours)              & \textbf{98.2} & 95.4          & \textbf{94.3} & \textbf{96.0}\\
\Xhline{1.5pt}
\end{tabular}
\end{center}
\end{table}

\begin{table}
\begin{center}
\caption{ Ablation experiments conducted for different settings on silhouette branch and skeleton branch. TS denotes the temporal stream. TSA and TC represent the temporal self-attention and temporal convolution respectively.}\label{tab:ablation2}
\vspace{+0.2cm}
\begin{tabular}{l|l|lll|l}
\Xhline{1.5pt}
\multicolumn{1}{c|}{Branch}  & \multicolumn{1}{c|}{Setting} & \multicolumn{1}{c}{NM} & \multicolumn{1}{c}{BG} & \multicolumn{1}{c}{CL} & \multicolumn{1}{|c}{Mean} \\
\cline{1-6}
\multirow{2}{*}{Silhouette} & w/o TS                      & 94.9                   & 89.7                   & 74.8                   & 86.5                     \\
                            & w/ TS                        & \textbf{96.3}          & \textbf{92.7}          & \textbf{81.0}          & \textbf{90.0}            \\
\cline{1-6}
\multirow{2}{*}{Skeleton}   & JSA-TSA                     & 74.8                   & 70.0                   & 65.6                   & 70.1                     \\
                            & JSA-TC                      & \textbf{92.3}          & \textbf{84.4}          & \textbf{78.9}          & \textbf{85.2}   \\
\Xhline{1.5pt}
\end{tabular}
\end{center}
\vspace{-0.2cm}
\end{table}

\subsection{Ablation Study}
Ablation experiments are conducted on CASIA-B to analyze different components of the proposed network, as shown in Table~\ref{tab:ablation1}. Besides, to validate the effectiveness of the unimodal structure, further experiments were conducted on the different settings in the silhouette branch and skeleton branch as shown in Table~\ref{tab:ablation2}.

\textbf{Effectiveness of the bimodal fusion}. The result a and b in Table~\ref{tab:ablation1} shows the averaged accuracies of the proposed silhouette branch and skeleton branch, respectively. The line c in Table~\ref{tab:ablation1} shows the performance achieved by combining the two branches. We can see that the two-branch network gets a better performance, which means that the complementary information contained in the skeleton and silhouette can help improve the generalization and feature representation capability of the gait recognition network.

\textbf{Effectiveness of the triple-branch architecture}. The comparison between c and e in Table~\ref{tab:ablation1} demonstrates the effectiveness of the proposed triple branch architecture, which utilizes the fusion branch to explore the mutual information between modalities and fully take advantage of the complementary of two modalities.

\textbf{Effectiveness of the aligning strategy}. To validate the effectiveness of the cross-modal token generator in the fusion branch, we apply the uniform partition (UP) strategy to segment the silhouette. The comparison between f and e in Table~\ref{tab:ablation1} reveals that the proposed aligning strategy and cross-modal learning method can effectively fuse the low-level feature from heterogeneous data, i.e., silhouette and skeleton.

\textbf{Structure of the silhouette branch}. The first 2 rows in Table~\ref{tab:ablation1} demonstrate the impact of the temporal stream in the silhouette branch. Without the temporal stream (w/o TS), the rank 1 accuracy declines by 3.5\%, which indicates that the MTA in the temporal stream is success capturing the motion pattern with different temporal granularity and constructing a complementary with the spatial stream.

\textbf{Structure of the silhouette branch}. To explore the temporal modeling method in the skeleton branch, we replace the temporal convolution with self-attention to model the joints dynamic change along the temporal dimension in the JSA-TC module. As shown in Table~\ref{tab:ablation2}, the results of JSA-TSA significantly decreased. One possible reason is that the movements of the joint are relatively simple and linear, while the self-attention mechanism is more suitable for capturing long-range dependencies which is less important for the skeleton.

\section{Conclusion}
In this work, our research focuses on improving the integration of multiple modes of gait data, through a novel triple branch network architecture called TriGait. We propose a two-stream network for the silhouette branch, which captures both static and motion information, and a JSA-TC module for the skeleton branch, which constructs the correlations between joints and capture dynamic information. Moreover, a novel cross-modal learning method is leveraged to align and fuse the two modalities from low-level features, enabling to fully exploit the complementary strengths of silhouette and skeleton. The proposed TriGait fusion network outperforms the state-of-the-art methods on both unimodal and bimodal gait recognition, which demonstrates the superiority and effectiveness of the triple branch fusion network. The ablation study provides further evidence of the effectiveness of each branch and the improved recognition performance by aligning and fusing the silhouette and skeleton via a tri-branch network.

This paper largely underlines the limit of the utility of using CASIA-B~\cite{yu2006framework} in respect of the newer and more challenging datasets. We look forward to the evaluation on other more demanding datasets, like 
 SUSTech\footnote{https://hid2023.iapr-tc4.org/} or GREW~\cite{zhu2021gait} in the future.

\section{Acknowledgment}
This work is funded by National Natural Science Foundation of China, grant number: 62002215. This work is funded by Shanghai Pujiang Program (No. 20PJ1404400).

{\small
\bibliographystyle{ieee}
\bibliography{egbib}

\begin{thebibliography}{10}\itemsep=-1pt

\bibitem{baltruvsaitis2018multimodal}
T.~Baltru{\v{s}}aitis, C.~Ahuja, and L.-P. Morency.
\newblock Multimodal machine learning: A survey and taxonomy.
\newblock {\em IEEE transactions on pattern analysis and machine intelligence},
  41(2):423--443, 2018.

\bibitem{castro2017evaluation}
F.~M. Castro, M.~J. Mar{\'\i}n-Jim{\'e}nez, N.~Guil, S.~L{\'o}pez-Tapia, and
  N.~P. de~la Blanca.
\newblock Evaluation of cnn architectures for gait recognition based on optical
  flow maps.
\newblock In {\em 2017 international conference of the biometrics special
  interest group (BIOSIG)}, pages 1--5. IEEE, 2017.

\bibitem{chao2019gaitset}
H.~Chao, Y.~He, J.~Zhang, and J.~Feng.
\newblock Gaitset: Regarding gait as a set for cross-view gait recognition.
\newblock In {\em Proceedings of the AAAI conference on artificial
  intelligence}, volume~33, pages 8126--8133, 2019.

\bibitem{connor2018biometric}
P.~Connor and A.~Ross.
\newblock Biometric recognition by gait: A survey of modalities and features.
\newblock {\em Computer vision and image understanding}, 167:1--27, 2018.

\bibitem{fan2020gaitpart}
C.~Fan, Y.~Peng, C.~Cao, X.~Liu, S.~Hou, J.~Chi, Y.~Huang, Q.~Li, and Z.~He.
\newblock Gaitpart: Temporal part-based model for gait recognition.
\newblock In {\em Proceedings of the IEEE/CVF conference on computer vision and
  pattern recognition}, pages 14225--14233, 2020.

\bibitem{hermans2017defense}
A.~Hermans, L.~Beyer, and B.~Leibe.
\newblock In defense of the triplet loss for person re-identification.
\newblock {\em arXiv preprint arXiv:1703.07737}, 2017.

\bibitem{huang2022gaitmstp}
B.~Huang, C.~Zhou, C.~Xu, and J.~Pan.
\newblock Gaitmstp: Multi-granularity spatio-temporal pyramid for gait
  recognition under complex covariation conditions.
\newblock In {\em 2022 IEEE International Joint Conference on Biometrics
  (IJCB)}, pages 1--8. IEEE, 2022.

\bibitem{lan2014multimedia}
Z.-z. Lan, L.~Bao, S.-I. Yu, W.~Liu, and A.~G. Hauptmann.
\newblock Multimedia classification and event detection using double fusion.
\newblock {\em Multimedia tools and applications}, 71:333--347, 2014.

\bibitem{li2023transgait}
G.~Li, L.~Guo, R.~Zhang, J.~Qian, and S.~Gao.
\newblock Transgait: Multimodal-based gait recognition with set transformer.
\newblock {\em Applied Intelligence}, 53(2):1535--1547, 2023.

\bibitem{li2020gait}
X.~Li, Y.~Makihara, C.~Xu, Y.~Yagi, and M.~Ren.
\newblock Gait recognition via semi-supervised disentangled representation
  learning to identity and covariate features.
\newblock In {\em Proceedings of the IEEE/CVF Conference on Computer Vision and
  Pattern Recognition}, pages 13309--13319, 2020.

\bibitem{liao2020model}
R.~Liao, S.~Yu, W.~An, and Y.~Huang.
\newblock A model-based gait recognition method with body pose and human prior
  knowledge.
\newblock {\em Pattern Recognition}, 98:107069, 2020.

\bibitem{lin2022gaitgl}
B.~Lin, S.~Zhang, M.~Wang, L.~Li, and X.~Yu.
\newblock Gaitgl: Learning discriminative global-local feature representations
  for gait recognition.
\newblock {\em arXiv preprint arXiv:2208.01380}, 2022.

\bibitem{lin2021gait}
B.~Lin, S.~Zhang, and X.~Yu.
\newblock Gait recognition via effective global-local feature representation
  and local temporal aggregation.
\newblock In {\em Proceedings of the IEEE/CVF International Conference on
  Computer Vision}, pages 14648--14656, 2021.

\bibitem{mao2020gait}
M.~Mao and Y.~Song.
\newblock Gait recognition based on 3d skeleton data and graph convolutional
  network.
\newblock In {\em 2020 IEEE International Joint Conference on Biometrics
  (IJCB)}, pages 1--8. IEEE, 2020.

\bibitem{peng2021learning}
Y.~Peng, K.~Ma, Y.~Zhang, and Z.~He.
\newblock Learning rich features for gait recognition by integrating skeletons
  and silhouettes.
\newblock {\em arXiv preprint arXiv:2110.13408}, 2021.

\bibitem{pinyoanuntapong2022gaitmixer}
E.~Pinyoanuntapong, A.~Ali, P.~Wang, M.~Lee, and C.~Chen.
\newblock Gaitmixer: skeleton-based gait representation learning via
  wide-spectrum multi-axial mixer.
\newblock {\em arXiv preprint arXiv:2210.15491}, 2022.

\bibitem{simonyan2014two}
K.~Simonyan and A.~Zisserman.
\newblock Two-stream convolutional networks for action recognition in videos.
\newblock {\em Advances in neural information processing systems}, 27, 2014.

\bibitem{sivapalan2011gait}
S.~Sivapalan, D.~Chen, S.~Denman, S.~Sridharan, and C.~Fookes.
\newblock Gait energy volumes and frontal gait recognition using depth images.
\newblock In {\em 2011 International Joint Conference on Biometrics (IJCB)},
  pages 1--6. IEEE, 2011.

\bibitem{song2020stronger}
Y.-F. Song, Z.~Zhang, C.~Shan, and L.~Wang.
\newblock Stronger, faster and more explainable: A graph convolutional baseline
  for skeleton-based action recognition.
\newblock In {\em proceedings of the 28th ACM international conference on
  multimedia}, pages 1625--1633, 2020.

\bibitem{sun2019deep}
K.~Sun, B.~Xiao, D.~Liu, and J.~Wang.
\newblock Deep high-resolution representation learning for human pose
  estimation.
\newblock In {\em Proceedings of the IEEE/CVF conference on computer vision and
  pattern recognition}, pages 5693--5703, 2019.

\bibitem{takemura2018multi}
N.~Takemura, Y.~Makihara, D.~Muramatsu, T.~Echigo, and Y.~Yagi.
\newblock Multi-view large population gait dataset and its performance
  evaluation for cross-view gait recognition.
\newblock {\em IPSJ transactions on Computer Vision and Applications},
  10:1--14, 2018.

\bibitem{teepe2022towards}
T.~Teepe, J.~Gilg, F.~Herzog, S.~H{\"o}rmann, and G.~Rigoll.
\newblock Towards a deeper understanding of skeleton-based gait recognition.
\newblock In {\em Proceedings of the IEEE/CVF Conference on Computer Vision and
  Pattern Recognition}, pages 1569--1577, 2022.

\bibitem{vaswani2017attention}
A.~Vaswani, N.~Shazeer, N.~Parmar, J.~Uszkoreit, L.~Jones, A.~N. Gomez,
  {\L}.~Kaiser, and I.~Polosukhin.
\newblock Attention is all you need.
\newblock {\em Advances in neural information processing systems}, 30, 2017.

\bibitem{wang2023combining}
L.~Wang, R.~Han, and W.~Feng.
\newblock Combining the silhouette and skeleton data for gait recognition.
\newblock In {\em ICASSP 2023-2023 IEEE International Conference on Acoustics,
  Speech and Signal Processing (ICASSP)}, pages 1--5. IEEE, 2023.

\bibitem{woo2018cbam}
S.~Woo, J.~Park, J.-Y. Lee, and I.~S. Kweon.
\newblock Cbam: Convolutional block attention module.
\newblock In {\em Proceedings of the European conference on computer vision
  (ECCV)}, pages 3--19, 2018.

\bibitem{yu2006framework}
S.~Yu, D.~Tan, and T.~Tan.
\newblock A framework for evaluating the effect of view angle, clothing and
  carrying condition on gait recognition.
\newblock In {\em 18th international conference on pattern recognition
  (ICPR'06)}, volume~4, pages 441--444. IEEE, 2006.

\bibitem{zhu2021gait}
Z.~Zhu, X.~Guo, T.~Yang, J.~Huang, J.~Deng, G.~Huang, D.~Du, J.~Lu, and
  J.~Zhou.
\newblock Gait recognition in the wild: A benchmark.
\newblock In {\em Proceedings of the IEEE/CVF international conference on
  computer vision}, pages 14789--14799, 2021.

\end{thebibliography}
}

\end{document}